\documentclass{article}
\usepackage{spconf,amsmath,graphicx}

\usepackage{enumitem}
\usepackage{booktabs}
\usepackage{amssymb}
\usepackage{url}
\setlist{nosep, leftmargin=14pt}
\let\oldthebibliography\thebibliography
\renewcommand{\thebibliography}[1]{%
  \oldthebibliography{#1}%
  \setlength{\itemsep}{0pt}%
  \setlength{\parsep}{0pt}%
  \setlength{\parskip}{0pt}%
}

\usepackage{mwe} 

\title{The Role of Recurrent Refinement in Medical Image Segmentation}
\name{Jo\~ao D.S. Marques$^{1}$, David Calhas$^{2}$, Arlindo L. Oliveira$^{1}$}
\address{$^{1}$INESC-ID / Instituto Superior T\'ecnico, Lisbon, Portugal\\
         $^{2}$Faculty of Computer and Information Science, University of Ljubljana}
\begin{document}
%
\maketitle
\begin{abstract}
While feed-forward segmentation networks perform well on large cohorts, they degrade in constrained clinical settings with few training samples and imaging artifacts. Human vision, by contrast, remains robust under these conditions, a property partly attributed to the recurrent and feedback connectivity of the visual cortex, which refines visual information over time. Inspired by this, we append lightweight recurrent refinement blocks to segmentation backbones, namely Conditional Random Fields (CRFs), modern Hopfield networks and convolutional GRUs, and study their behaviour as the number of iterations increases at test time (including beyond the training horizon). Evaluated on two open-source datasets, Kvasir-SEG and ARCADE, our recurrent blocks add less than 1.5\% parameters and yield gains on clean data (up to 1.4 pp) and under noise (up to 17 pp) while reducing HD95.

\end{abstract}

\section{Introduction}
\label{sec:intro}

Semantic segmentation is a critical task in clinical workflows, as it provides spatial delineations essential for surgical navigation, lesion tracking, and morphology quantification \cite{survey_segmentation_medicine}. Over the past decade, architectures ranging from standard convolutional U-Nets \cite{unet} to hierarchical Vision Transformers \cite{segformer} and medical foundation models like MedSAM \cite{MedSAM} have demonstrated exceptional performance, particularly when trained on very large benchmarks. However, real-world clinical settings rarely involve large scale cohorts, as most accessible medical datasets are severely constrained by small training sets, acquisition artifacts and label noise \cite{medical_datasets_are_noisy}. As a consequence, these models struggle in many scenarios and clinical deployment requires the capacity of extracting the most utility from such conditions.

In contrast to artificial networks, the primate visual system exhibits remarkable robustness in noisy and low-data regimes \cite{resistencia_humana_ruido,recurrence_ventral_stream}. While architectures like the U-Net try to mirror hierarchical receptive fields, these models remain strictly feed-forward, mapping inputs to predictions in a single static computational pass without the capability to re-evaluate intermediate features. Instead, the human visual cortex functions as an active dynamical system with pervasive recurrent and top-down feedback pathways, refining decisions over time to resolve spatial ambiguity and sensory noise \cite{describing_visual_cortex,recurrent_neural_networks_explain_things,human_like_metrics_recurrence_models}. Therefore, incorporating recurrent feedback mechanisms into segmentation models offers an intuitive approach to equip them with iterative, dynamic error correction mechanisms.

Prior studies have used recurrent computation through various paradigms, such as Deep Equilibrium Models (DEQ) that enforce convergence toward stable fixed points \cite{deep_equilibrium_models}, unrolled recurrent convolutions within U-Net blocks \cite{Recurrent_Resunet_medical_segmentation}. Similarly, formulating Dense Conditional Random Fields (CRF) with Gaussian edge potentials as Recurrent Neural Networks (CRF-as-RNN)~\cite{crf_field_filter,CRF_as_RNN} demonstrated that mean-field message passing can iteratively sharpen boundary beliefs. In parallel, Modern Hopfield Networks \cite{hopfield_is_all_you_need,energy_transformer} established a formal bridge between associative memory retrieval and attention, proving that spatial tokens can converge toward stable energy attractors. Beyond energy-based and probabilistic frameworks, RAFT \cite{raft_architecture} introduced an alternative by leveraging Convolutional Gated Recurrent Units (ConvGRUs) \cite{convgru} to keep a spatial memory state and iteratively emit bounded, residual corrections over time. Nevertheless, existing recurrent segmentation methods predominantly evaluate performance under a fixed, small number of unrolled iterations ($T$), leaving open whether segmentation quality improves and remains stable as $T$ increases at test time, including beyond the training horizon.

In this work, we investigate the role of recurrent blocks plugged into a pre-trained segmentation backbone. Specifically, we evaluate CRF \cite{CRF_as_RNN}, modern Hopfield networks \cite{hopfield_is_all_you_need} and ConvGRU \cite{convgru} integrated at the output of a segmentation model. We benchmark these architectures across varying iterations $T$ at test time under synthetic corruptions and few-shot constraints on gastrointestinal polyp segmentation (Kvasir-SEG \cite{dataset_kvasir}) and coronary angiography (ARCADE \cite{dataset_ARCADE}). All code available at \url{https://github.com/joaodsmarques/recurrencehelpsegmentation}.

\section{Methodology}
\label{sec:methods}
\subsection{The Datasets}
This study is evaluated on two medical image segmentation benchmarks of contrasting complexity: Kvasir-SEG \cite{dataset_kvasir} and the syntax split of ARCADE \cite{dataset_ARCADE}. All images and corresponding ground-truth masks were resized to a fixed resolution of $512 \times 512$ pixels. For Kvasir-SEG, the cohort was partitioned into an 80\%–10\%–10\% split from a total of 1000 pairs using a fixed random seed; the held-out test set remained isolated across experiments, while validation partitions varied across folds. For ARCADE, we adopted the official benchmark partition comprising 1000 training, 200 validation, and 300 test images. These datasets provide complementary evaluation settings: Kvasir-SEG represents a localized lesion delineation task with different textures, whereas ARCADE poses a difficult tubular tracking scenario composed of thin scale vascular geometry, connectivity constraints and high noise. Both tasks were framed as single target binary segmentation: polyp identification in Kvasir-SEG and coronary vessel delineation in ARCADE. 

\subsection{Recurrent Refinement Framework}
\label{sec:recurrent_framework}
Let $f_\theta$ denote a feed-forward backbone (U-Net for example) that maps an input image $\mathbf{x} \in \mathbb{R}^{3 \times H \times W}$ to unary logits $\mathbf{U} = f_\theta(\mathbf{x}) \in \mathbb{R}^{C \times H \times W}$ \cite{crf_field_filter}. A shared recurrent operator $g_\phi$, where $g_\phi : \mathbb{R}^{C \times H \times W} \times \mathcal{C} \to \mathbb{R}^{C \times H \times W}$ is a parameterized recurrent correction operator that computes a bounded residual update $\Delta^t = g_\phi(\mathbf{L}^{t-1}, \mathbf{c})$ conditioned on optional feature/image context $\mathbf{c}$, is executed for $T$ iterations starting from $\mathbf{L}^{0} = \mathbf{U}$:
\begin{equation}
\mathbf{L}^{t} = \mathbf{U} + g_\phi(\mathbf{L}^{t-1}, \mathbf{c}), \quad t = 1, \dots, T,
\label{eq:rec_simple}
\end{equation}
where $\mathbf{c}$ denotes optional context and $\mathbf{Q}^{t} = \mathrm{softmax}(\mathbf{L}^{t})$. Here, $\mathbf{U}$ works as a persistent skip-connection anchoring every step to the initial prediction and preventing an unbounded divergence as $T$ scales. By re-injecting $\mathbf{U}$ at every step, the operator predicts bounded residual corrections relative to static feed-forward evidence rather than accumulating unbounded state drift across successive unrolled passes. All residual projection heads in $g_\phi$ are initialized to zero, ensuring that the model at $t=0$ is strictly identical to the base network. To encourage that increasing the iteration count $T$ progressively refines the prediction rather than destabilizing it, we adopt a step-weighted supervision strategy inspired by RAFT~\cite{raft_architecture}, using linearly increasing rather than exponential weights. By supervising intermediate outputs across all recurrent passes, the loss penalizes later iterations more heavily:
\begin{equation}
\mathcal{L} = \frac{\sum_{t=t_0}^{T} (t+1)\,\ell(\mathbf{L}^{t}, \mathbf{y})}{\sum_{t=t_0}^{T} (t+1)},
\label{eq:deep_supervision}
\end{equation}
where $\ell = 0.6\,\mathcal{L}_{\text{Dice}} + 0.4\,\mathcal{L}_{\text{CE}}$, and $t_0 = 0$. This weighting lets early iterations establish coarse structure while placing more emphasis on the accuracy of later steps, favouring fine-grained corrections. The chosen composed loss function balances region overlap with pixel-level classification: the Dice component addresses severe foreground-background imbalance inherent to thin vessels and localized lesions, while Cross-Entropy provides smooth, well-calibrated gradients that stabilize backpropagation across recurrent iterations.

\begin{figure}
    \centering
    \includegraphics[width=0.95\linewidth]{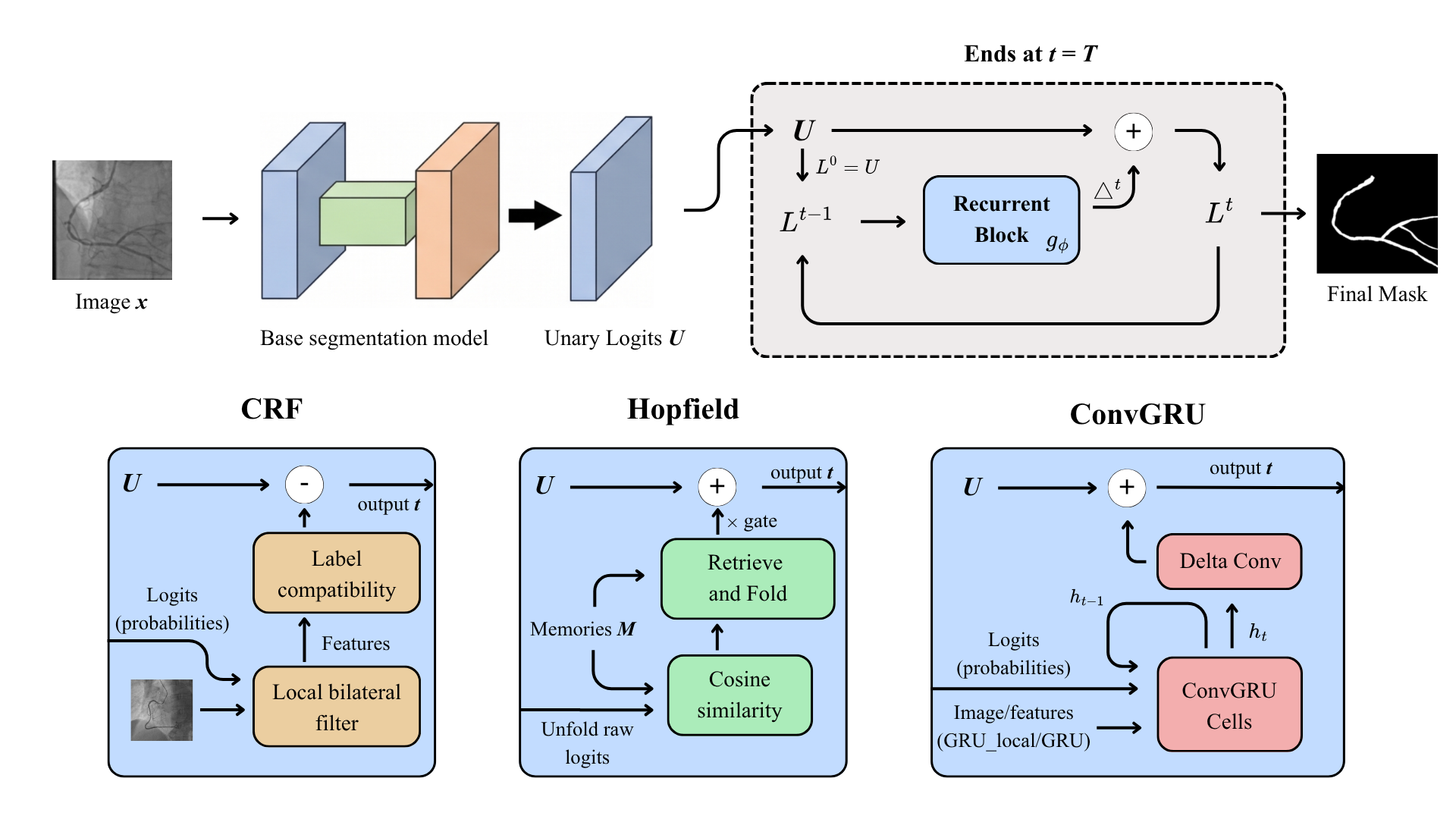}
    \caption{Overview of the recurrent refinement framework.}
    \label{fig:pipeline}
\end{figure}

\subsection{Recurrent blocks}

The full pipeline using each of the chosen recurrent blocks is shown in Figure~\ref{fig:pipeline}.

\textbf{CRF:} Following CRF-as-RNN \cite{CRF_as_RNN}, $g_\phi$ computes local bilateral filtering over a $7 \times 7$ window on downsampled logits and routes the resulting messages through a learnable $1 \times 1$ compatibility matrix. This explicitly enforces spatial and color-dependent label consistency. The detailed mean-field derivation can be found in Zheng et al. work \cite{CRF_as_RNN}.

\textbf{Hopfield Network:} The block $g_\phi$ unfolds centered probabilities $\mathbf{Q}^{t-1} - 1/C$, with $C = 2$ (background and target), into overlapping $16 \times 16$ patches (stride = 8), retrieves prototypical shape patterns from a learned dictionary of 512 memories via continuous modern Hopfield attention \cite{hopfield_is_all_you_need}, and folds them back into logit space. The associative memory functions as a data driven geometric prior.

\textbf{ConvGRU:} Parameterized as a convolutional GRU \cite{convgru,raft_architecture}, $g_\phi$ updates an internal spatial hidden state $\mathbf{h}^t$ conditioned on decoder feature context $\mathbf{c}$ and current label probabilities. At each iteration, it emits a bounded residual correction $\Delta^t$ via a convolutional head initialized at 0, that expands its effective receptive field with every step.

\textbf{ConvGRU-local:} To emulate an unconstrained, non-linear CRF without semantic shortcuts, ConvGRU-local restricts the context $\mathbf{c}$ exclusively to raw input image patches. In addition, step-0 unary supervision is disabled ($t_0 = 1$), enabling the feed-forward backbone to provide coarse proposals that are systematically refined through local propagation.

\begin{table}[t]
\centering
\caption{Clean test set results for $T$ recurrent steps at test time: mean (std) across 5 training seeds. Dice and precision in \%, HD95 in pixels. Best values in bold.}
\label{tab:blocks}
\setlength{\tabcolsep}{3pt}
\footnotesize
\begin{tabular}{llcccccc}
\toprule
 & & \multicolumn{3}{c}{ARCADE} & \multicolumn{3}{c}{Kvasir-SEG} \\
\cmidrule(lr){3-5}\cmidrule(lr){6-8}
Model & $T$ & Dice$\uparrow$ & Prec.$\uparrow$ & HD95$\downarrow$ & Dice$\uparrow$ & Prec.$\uparrow$ & HD95$\downarrow$ \\
\midrule
U-Net & -- & 79.7{\scriptsize(.2)} & 82.3{\scriptsize(.2)} & 62.4{\scriptsize(1.8)} & 89.1{\scriptsize(.4)} & 90.1{\scriptsize(2.6)} & 42.2{\scriptsize(.9)} \\
\midrule
CRF & 5 & 79.3{\scriptsize(.2)} & 82.7{\scriptsize(1.3)} & 62.7{\scriptsize(1.5)} & 89.0{\scriptsize(.7)} & 87.8{\scriptsize(2.9)} & 40.7{\scriptsize(2.4)} \\
 & 20 & 79.5{\scriptsize(.2)} & 83.6{\scriptsize(1.3)} & 64.4{\scriptsize(1.8)} & 89.0{\scriptsize(.7)} & 89.3{\scriptsize(2.4)} & 41.1{\scriptsize(3.1)} \\
\midrule
Hopf & 5 & \textbf{80.9}{\scriptsize(.2)} & 83.9{\scriptsize(.3)} & \textbf{60.0}{\scriptsize(1.4)} & 89.8{\scriptsize(1.7)} & 90.0{\scriptsize(1.9)} & 39.7{\scriptsize(3.2)} \\
 & 20 & 80.6{\scriptsize(.2)} & 84.3{\scriptsize(.3)} & 60.2{\scriptsize(1.3)} & 89.9{\scriptsize(1.7)} & 90.1{\scriptsize(1.9)} & 39.6{\scriptsize(3.9)} \\
\midrule
GRU & 5 & 79.5{\scriptsize(.3)} & \textbf{85.8}{\scriptsize(.9)} & 61.0{\scriptsize(.4)} & 89.8{\scriptsize(1.3)} & 91.4{\scriptsize(1.7)} & \textbf{37.7}{\scriptsize(6.5)} \\
 & 20 & 79.4{\scriptsize(.3)} & 85.7{\scriptsize(1.2)} & 61.3{\scriptsize(.7)} & 89.5{\scriptsize(1.3)} & 91.3{\scriptsize(1.4)} & 38.8{\scriptsize(8.1)} \\
\midrule
GRU-L & 5 & 80.3{\scriptsize(.3)} & 83.2{\scriptsize(.7)} & 60.8{\scriptsize(2.7)} & \textbf{90.5}{\scriptsize(1.3)} & 89.8{\scriptsize(1.3)} & 39.4{\scriptsize(4.8)} \\
 & 20 & 80.1{\scriptsize(.3)} & 83.4{\scriptsize(.5)} & 61.6{\scriptsize(2.3)} & 90.1{\scriptsize(1.3)} & \textbf{91.6}{\scriptsize(1.2)} & 39.6{\scriptsize(4.0)} \\
\bottomrule
\end{tabular}
\end{table}

\begin{figure}
    \centering
    \includegraphics[width=0.95\linewidth]{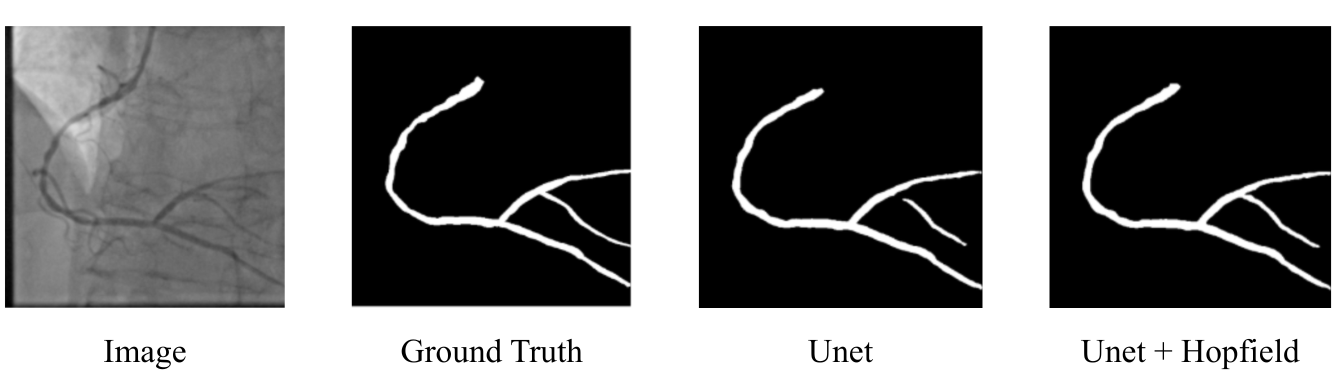}
    \caption{Recurrence visual effect on masks.}
    \label{fig:visual results}
\end{figure}

\section{Experiments and Results}
\label{sec:pagestyle}
We structure our evaluation around three setups: 
(i) a clean data benchmark, evaluating whether appending recurrent refinement blocks improves performance over the feed-forward; (ii) a low data regime, where models are trained on progressively smaller subsets of the training split, while keeping validation and test sets fixed to assess data efficiency; (iii) an input corruption stress test, where we add synthetic noise profiles (Gaussian, Poisson, and blur) to measure architectural resilience against degraded visual inputs.

\subsection{Implementation details}
For the main backbone, we chose a U-Net featuring an EfficientNet-B4 encoder \cite{efficientnet} initialized with ImageNet pre-trained weights as it provides a strong trade-off between performance and parameter efficiency. Hyperparameters were optimized using Optuna across 100 trials with median pruning over a 30 epoch limit, maximizing the validation Dice score. To ensure a fair comparison, all recurrent variants were initialized directly from the ImageNet pre-trained U-Net rather than the one fine-tuned with Optuna. 

The learning rates were set to approximately $2 \times 10^{-4}$ for the baseline U-Net, $3 \times 10^{-4}$ for the CRF, and $5 \times 10^{-4}$ for both Hopfield and ConvGRU configurations. The parameters associated with the recurrent blocks were assigned a learning rate scaled by a factor of 3 relative to the backbone, with weight decay disabled. All models were trained for 50 epochs using a batch size of 8, weight decay of $1 \times 10^{-4}$ (backbone), and a cosine learning rate scheduler across four 32\,GB NVIDIA V100 GPUs installed on a Dell PowerEdge C4140 server. The quality of each architecture is evaluated via the Dice similarity coefficient and precision for regional accuracy, along with the 95th-percentile Hausdorff Distance (HD95) to quantify boundary preservation and topological connectivity, which is particularly important for the ARCADE dataset. All recurrent blocks are trained with deep supervision: the loss is applied to the output of every step $t$ with weight $t{+}1$, so later steps are weighted more and each step is encouraged to improve on the previous one (for ConvGRU-local, step 0 is not supervised, so the base network only provides a proposal). The CRF and Hopfield blocks are trained with a fixed $T{=}5$. The ConvGRU blocks are trained with $T$ drawn uniformly from $\{1,\dots,8\}$ for every batch, so that the output is usable after any number of steps. The Hopfield update converges mostly in the first steps, and training it with $T\in\{1,3,5\}$ presented differences within seed variability, so we keep $T{=}5$ for consistency with the CRF. All experiments were run on 5 different seeds and k-fold validation with 5 different folds for Kvasir dataset.

\begin{figure}
    \centering
    \includegraphics[width=0.95\linewidth]{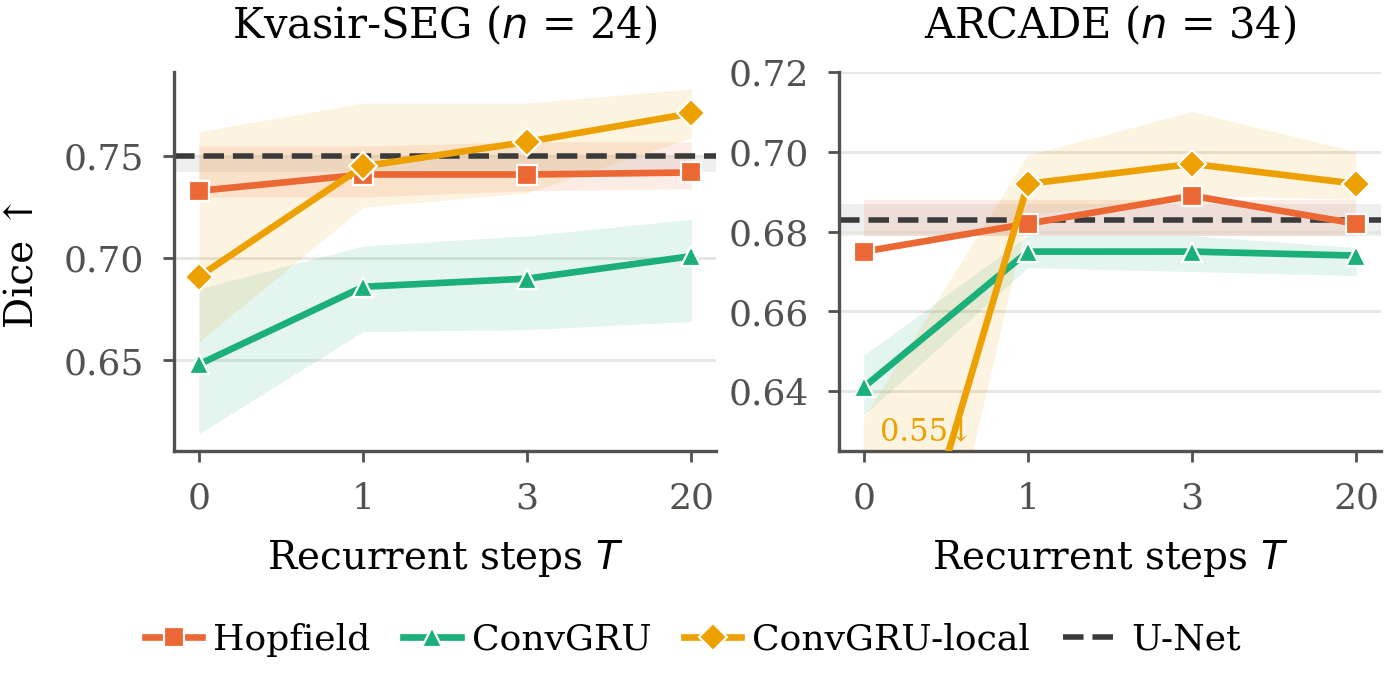}
    \caption{How iterations influence the Dice score in limited samples settings.}
    \label{fig:iterations_study}
\end{figure}

\subsection{Clean data benchmark}

Table~\ref{tab:blocks} reports results on the clean test sets. Because the intermediate outputs are supervised during training (deep supervision), all blocks produce valid predictions for any $T$ at test time, including beyond the training horizon; the only exception is step 0 of ConvGRU-local, which is not supervised. Statistical significance was assessed with a two-sided paired Wilcoxon signed-rank test on per-image metrics (averaged over training seeds), with Holm correction for multiple comparisons.
On clean data, gains over the feed-forward baseline are modest. The best recurrent variant improves Dice by +1.2\,pp on ARCADE (Hopfield, $T{=}5$) and +1.4 pp on Kvasir-SEG (ConvGRU-local, $T{=}5$). The largest HD95 reductions are 2.4 px on ARCADE (Hopfield, $T{=}5$) and 4.5 px on Kvasir-SEG (ConvGRU, $T{=}5$), although this last one is smaller than its standard deviation across seeds ($\pm$6.5 px).

The blocks mainly shift the operating point towards precision. On ARCADE, all blocks increase precision over the U-Net ($p<0.05$), with ConvGRU showing the largest gain (+3.5 pp). The CRF does not improve Dice on either dataset, but it exhibits the clearest step-wise behaviour: on ARCADE, precision rises from 79.4\% at $T{=}1$ to 82.7\% at $T{=}5$ and 83.6\% at $T{=}20$ ($p<0.001$,
consistent in 5/5 seeds), with no significant change in Dice. On Kvasir-SEG, CRF precision also increases with $T$ (87.8\% to 89.3\%) but remains below the baseline. Mean-field message passing therefore acts as a refinement that suppresses low confidence foreground pixels. Extending inference from $T{=}5$ to $T{=}20$ changes Dice by at most 0.5 pp for every block. 

\subsection{Limited samples study}

Figure~\ref{fig:iterations_study} shows Dice as a function of T for models trained on n=24 (Kvasir-SEG) and n=34 (ARCADE) images, about 3\% of the respective training sets. ConvGRU-local starts below the baseline at T=0 because its step-0 output is unsupervised. However, this block improves with each iteration and surpasses the U-Net on both datasets (at T=3 in Kvasir-SEG and T=1 at ARCADE). The Hopfield block remains near the baseline and ConvGRU, although it improves with iterations, never surpasses the baseline. Variance across seeds is high in this regime and with larger subsets the gap between the best recurrent blocks and the baseline shrinks. We therefore read these results as suggestive rather than conclusive: iterative refinement can partially compensate for scarce data, but the benefit depends on the block and diminishes as data increase.
\subsection{Robustness to input corruption}

\begin{figure}
    \centering
    \includegraphics[width=0.95\linewidth, trim={9.5cm 0.5cm 9.5cm 0.4cm}, clip]{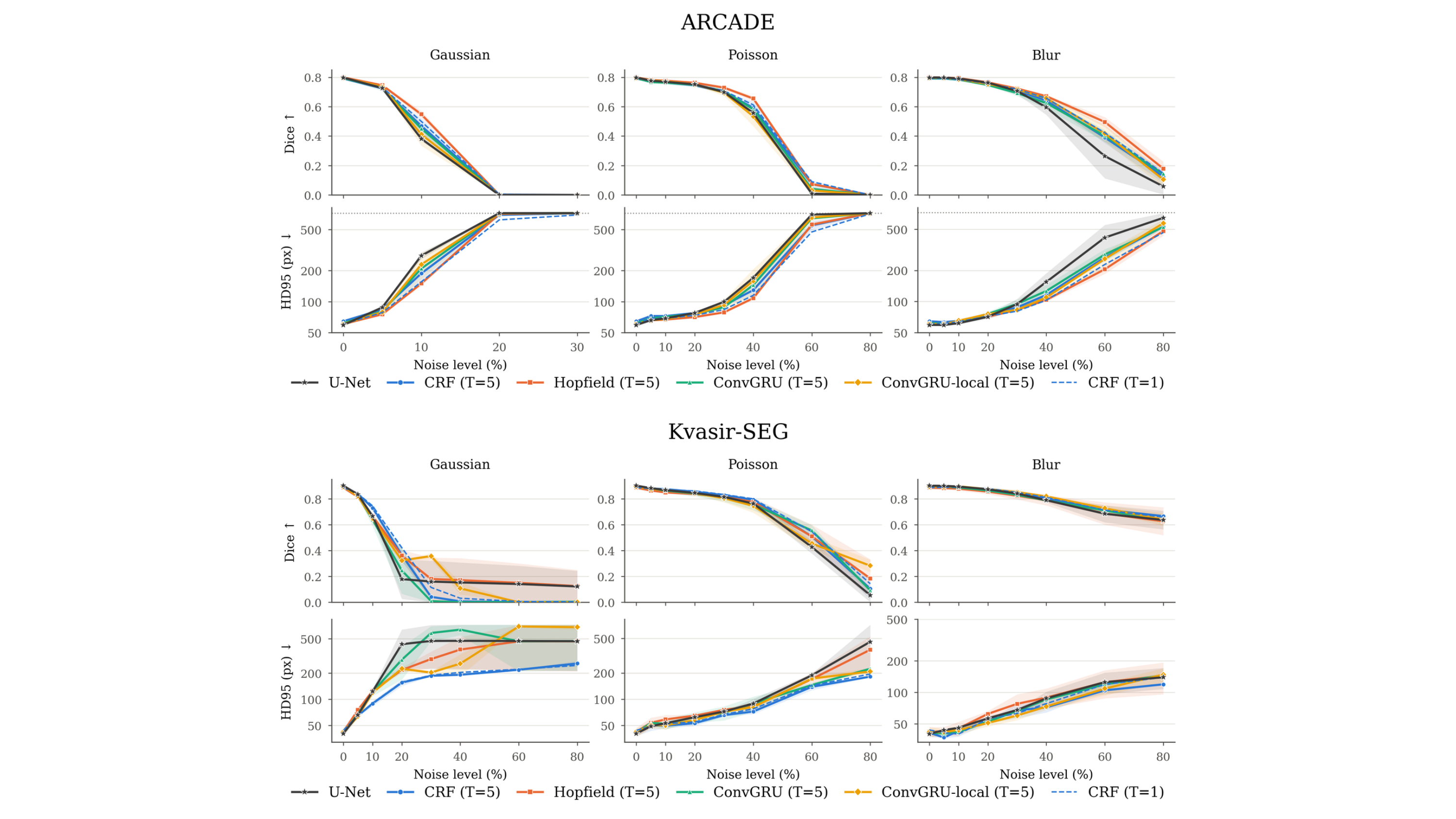}
    \caption{Noise robustness test in ARCADE and Kvasir-SEG.}
    \label{fig:noise_study}
\end{figure}

All models were trained on clean images and evaluated under Gaussian noise, Poisson noise, and Gaussian blur of increasing severity and results were reported in Figure~\ref{fig:noise_study}. On ARCADE, the Hopfield block (T=5) is the most robust model under all three noise types: at Gaussian 10\%, Poisson 40\% and blur 40\% it improves Dice by +17 pp, +10 pp and +8 pp and reduces HD95 by 129, 62 and 51 px over the U-Net ($p<0.001$). At high severities all models collapse, since no refinement can recover structure absent from the unary predictions. The pattern of larger gains in HD95 than in Dice suggests that the blocks mainly remove distant false positives.

\subsection{Compute cost}
To confirm that performance gains come from recurrence rather than increased model size, we report the number of trainable parameters, inference GFLOPs, and peak GPU training memory in Table~\ref{tab:cost}. The parameter overhead introduced by recurrence is marginal: both ConvGRU and Hopfield add less than 1.5\% capacity, while the CRF adds only six trainable parameters. Nevertheless, per-step costs differ substantially: while the CRF adds only 0.1 GFLOPs per iteration, ConvGRU adds 20.8 GFLOPs, so at $T{=}8$ its inference cost is roughly $5\times$ that of the base U-Net.

\begin{table}[t]
\centering
\caption{Cost of each block on our setting. We consider
GFLOPs per image as a function of the number of recurrent steps $T$ and memory as the peak GPU memory of one training step (batch 8, fp32, AdamW) at the training $T$ (5 for CRF and Hopfield, 8 for ConvGRU).}
\label{tab:cost}
\setlength{\tabcolsep}{4pt}
\footnotesize
\begin{tabular}{lccc}
\toprule
Model & Params (M) & GFLOPs & Mem. (GiB) \\
\midrule
U-Net           & 20.23 & 36.9             & 11.0 \\
+ CRF           & 20.23 (+0.0\%) &  36.9 + 0.1T & 13.8 (+25\%) \\
+ Hopfield      & 20.49(+1.3\%) & 36.9 + 4.2T     & 12.4 (+13\%) \\
+ ConvGRU       & 20.40(+0.8\%) & 38.1 + 20.8T    & 16.3 (+48\%) \\
\bottomrule
\end{tabular}
\end{table}

\subsection{Other backbones}
With a SegFormer (MiT-B2)~\cite{segformer} backbone (5 seeds), the CRF behaviour replicates: precision increases with T on both datasets (+4 pp, $p < 0.001$), and the CRF improves Dice and HD95 over the backbone under all three corruptions on ARCADE ($p < 0.02$). However, the best block under noise is backbone-dependent (CRF for SegFormer, Hopfield for U-Net). We also observed that when the base model is weak, recurrence barely changes the output, suggesting that the blocks refine the unary prediction rather than correct it.

\section{Conclusion}
We evaluated lightweight recurrent refinement blocks across test-time iteration counts. On clean data, gains are small. Nevertheless, under input corruption the blocks generally reduce HD95, mainly by suppressing spurious predictions. Predictions remain stable beyond the training horizon (Dice changes by at most 0.5 pp from T=5 to T=20), which we attribute to anchoring each step to the unary logits. Although the gains are modest, the impact on segmentation quality can be meaningful as presented by Figure~\ref{fig:visual results}. 

\section{COMPLIANCE WITH ETHICAL STANDARDS}
This research study was conducted using open-access human subject data from the Kvasir-SEG \cite{dataset_kvasir} and ARCADE \cite{dataset_ARCADE} benchmarks. Ethical approval was not required for the use of these anonymized, publicly available datasets, as confirmed by their respective licenses.

\section{Acknowledgments}
\label{sec:acknowledgments}
This work was supported by national funds through Fundação para a Ciência e a Tecnologia, I.P. (FCT) under projects UID/50021/2025 (\url{http://doi.org/10.54499/UID/50021/2025}) and UID/PRR/50021/2025 (\url{http://doi.org/10.54499/UID/PRR/50021/2025}).

\bibliographystyle{IEEEbib}
\bibliography{refs}

\end{document}